\documentclass[letterpaper]{article} % DO NOT CHANGE THIS
\usepackage{aaai2026}  % DO NOT CHANGE THIS
\usepackage{times}  % DO NOT CHANGE THIS
\usepackage{helvet}  % DO NOT CHANGE THIS
\usepackage{courier}  % DO NOT CHANGE THIS
\usepackage[hyphens]{url}  % DO NOT CHANGE THIS
\usepackage{graphicx} % DO NOT CHANGE THIS
\usepackage{natbib}  % DO NOT CHANGE THIS AND DO NOT ADD ANY OPTIONS TO IT
\usepackage{caption} % DO NOT CHANGE THIS AND DO NOT ADD ANY OPTIONS TO IT
\usepackage{algorithm}
\usepackage[table, dvipsnames]{xcolor}
\usepackage{graphicx} % Required for inserting images
\usepackage{booktabs}
\usepackage{xcolor}
\usepackage{amssymb}
\usepackage{newtxtext,newtxmath}
\usepackage{mathrsfs}
\usepackage{amsmath}
\usepackage{multirow}
\usepackage{soul}
\usepackage{color} 
\usepackage{algpseudocode}

\usepackage{newfloat}
\usepackage{listings}
\DeclareCaptionStyle{ruled}{labelfont=normalfont,labelsep=colon,strut=off} % DO NOT CHANGE THIS
\floatstyle{ruled}
\newfloat{listing}{tb}{lst}{}
\floatname{listing}{Listing}
\title{PANDA - Patch And Distribution-Aware Augmentation\\ for Long-Tailed Exemplar-Free Continual Learning}
\author {
    Siddeshwar Raghavan\textsuperscript{\rm 1},
    Jiangpeng He\textsuperscript{\rm 2,}\thanks{Corresponding Author, Project Lead},
    Fengqing Zhu\textsuperscript{\rm 1}
}
\affiliations {
    \textsuperscript{\rm 1}Department of Electrical and Computer Engineering, Purdue University, West Lafayette, USA\\
    \textsuperscript{\rm 2}Department of Computer Science, Indiana University, Bloomington, USA\\
    raghav12@purdue.edu, jhe2@iu.edu, zhu0@purdue.edu
}
\begin{document}

\maketitle

\begin{abstract}
Exemplar-Free Continual Learning (EFCL) restricts the storage of previous task data and is highly susceptible to catastrophic forgetting. While pre-trained models (PTMs) are increasingly leveraged for EFCL, existing methods often overlook the inherent imbalance of real-world data distributions. We discovered that real-world data streams commonly exhibit dual-level imbalances, dataset-level distributions combined with extreme or reversed skews within individual tasks, creating both intra-task and inter-task disparities that hinder effective learning and generalization.
To address these challenges, we propose PANDA, a Patch-and-Distribution-Aware Augmentation framework that integrates seamlessly with existing PTM-based EFCL methods. PANDA amplifies low-frequency classes by using a CLIP encoder to identify representative regions and transplanting those into frequent-class samples within each task. Furthermore, PANDA incorporates an adaptive balancing strategy that leverages prior task distributions to smooth inter-task imbalances, reducing the overall gap between average samples across tasks and enabling fairer learning with frozen PTMs. Extensive experiments and ablation studies demonstrate PANDA's capability to work with existing PTM-based CL methods, improving accuracy and reducing catastrophic forgetting.
\end{abstract}

\begin{links}
    \link{Code}{https://gitlab.com/viper-purdue/panda}
    % \link{Extended version}{https://aaai.org/example/extended-version}
\end{links}

% ---------------------------------------------------------------------
\section{Introduction}

Continual Learning systems have made remarkable strides in overcoming catastrophic forgetting~\cite{cf_1, cf_2, cf_3} and improving learning with growing streams of data. Despite these advancements, a large number of methods still assume perfectly balanced tasks with uniform class distributions~\cite{CIL_survey, comprehensive_CL_survey, CL_survey}. In reality, data streams are both sensitive and unconstrained, and they often follow long-tailed distributions at two levels. Globally, certain classes dominate while others are rare; within a task, this skew can be more extreme or temporarily reversed. This type of two level imbalance remains underexplored in continual learning, resulting in a gap between benchmark studies and real-world applications. For instance, camera traps may log thousands of deer and rabbits, but only a few predators. During a migration window, deer can dominate a week’s footage. In medical imaging, pneumonia is the most common condition, while pneumoconiosis appears sporadically and can occasionally overtake pneumonia as the most common case.

Data sensitivity and the high cost of storage make exemplar-based continual learning methods impractical in many real‑world settings. Historically, exemplar‑free~\cite{LWF, EWC, smith2023closerlookrehearsalfreecontinual} approaches could not match the performance of exemplar‑replay~\cite{ER} techniques. However, this has changed with the recent advancement of Pre-Trained Models (PTM) trained on extensive datasets, which has inspired a growing number of continual learning techniques that benefit from their rich feature representations~\cite{he2025cl, fecam, slca, CIL_PTM_survey, zhou2024revisitingclassincrementallearningpretrained, ranpac, L2P, codaprompt, dualprompt, simplecil_adam, APER, MOS}. Despite these advances, Exemplar-Free Continual Learning (EFCL) is still in its early stages when it comes to handling two-level imbalances. Prior work on imbalanced Long-Tailed-EFCL(LT-EFCL)~\cite{APART, DAP} addresses only what we term single‑level imbalance (SLI), designed to tackle a global long-tailed distribution. 

To address this, we first~\textbf{ formalize the Dual Level Imbalance (DLI)} setting by introducing task-specific imbalances that deviate from the dataset level distribution. Then, to handle both single-level imbalance (SLI) and our proposed DLI in PTM-based EFCL frameworks, we present PANDA, a Patch-and-Distribution-Aware oversampling module that integrates into any existing EFCL method. PANDA combines two complementary mechanisms: (1)\textbf{Intra‑task balancing} (within task), using a frozen CLIP encoder to selectively transfer semantic-rich patches from tail class samples to head class samples to equalize distribution.  
(2) For \textbf{Inter‑task smoothening} we blend the previous task’s minima and maxima with the current task via a learnable $\beta$ to calibrate classifier thresholds.

We evaluate PANDA across both single and dual level imbalance settings, showcasing the improvement in accuracy and reduced forgetting. Our contributions are as follows:

\begin{enumerate}
    \item We formalize the Dual Level Imbalance (DLI) setting to inject task specific imbalance that deviates from the overall dataset distribution. 
    \item We propose PANDA, an integratable  patch and distribution aware oversampling module that (a) transfers CLIP-identified patches from rare to common classes to balance intra-task distributions and (b)  smoothening inter-task distribution shifts via a learnable $\beta$, reducing classifier bias without storing past data.
    \item We integrate PANDA into existing PTM EFCL frameworks and demonstrate significant improvements in accuracy and mitigate catastrophic forgetting, highlighting its broad applicability and impact.
\end{enumerate}
% ---------------------------------------------------------------------
\section{Related Work}
\label{sec:related_work}

\subsection{Continual learning categories}

Traditionally, continual learning has primarily focused on deep learning models trained from scratch~\cite{CIL_survey, CL_survey, he_ocl}, categorized into replay-based methods, regularization techniques, and parameter-isolation approaches. \textbf{Replay-based} methods~\cite{icarl, ER, selectiveER, CL_episod, ocl_realworld} utilize a memory buffer to retain a subset of past data samples, enabling knowledge rehearsal to mitigate forgetting of previously learned classes. \textbf{Regularization} methods~\cite{LWF, EWC, IMM, lessforget} incorporate additional penalty terms into the loss function to preserve prior task information while effectively learning new data. In contrast, \textbf{parameter-isolation} techniques~\cite{Serr2018OvercomingCF, packnet, piggyback, progNN} assign separate model parameters to different tasks, preventing interference and mitigating forgetting. While most of these approaches focus on balanced settings, recent research has begun exploring \textbf{imbalanced} and \textbf{long-tailed} scenarios with replay strategies~\cite{he2023long, liu2022long, ranibowmemory, sraghavan2024delta, ocl_imbalance, DGR}. 

\subsection{Exemplar Free Continual Learning with PTM}

\textbf{Continual learning with PTMs}~\cite{CIL_PTM_survey} has demonstrated greater resilience to forgetting and improved performance compared to models trained from scratch, even when they're exemplar-free in nature, making them a promising direction for more efficient continual learning systems. Continual learning with pre-trained models can be broadly categorized into two main approaches: \textbf{Prompt-based} methods~\cite{L2P, codaprompt, dualprompt, DAP} and \textbf{Representation-based} methods~\cite{simplecil_adam, ranpac, ease, APART, cofima, slca, fecam, APER, MOS}. Prompt-based methods utilize lightweight, trainable parameters (prompts) to guide the model in learning task-specific image samples. These prompts are attached to the input alongside image patches, helping the model adapt efficiently to new tasks while leveraging pre-trained knowledge. Representation-based methods leverage pre-trained knowledge by keeping the backbone completely frozen while replacing classification weights with prototypes, using a nearest mean classifier for classification or adapters or a mix of these strategies.

\subsection{Data augmentation methods}
In the context of long-tailed learning and imbalanced datasets, oversampling tail-end classes is a common technique, but focusing on the semantic information is necessary to aid in the augmentation. The Cutout~\cite{cutout}
technique randomly removes regions from the image, while Mixup~\cite{mixup} creates new samples by interpolating two images in the dataset. In contrast, CutMix~\cite{cutmix} cuts a patch from one image and pastes it onto another, while also adjusting the labels proportionally. A recent data augmentation technique for long-tailed learning employs contrastive learning to generate semantically consistent data, thereby addressing the imbalance~\cite{concutmix}. However, none of these methods are inherently designed for continual learning, where the entire sample distribution is unknown. While balancing task distributions can help mitigate imbalance, the inherent distribution shifts across different tasks can still lead to a biased classifier. Our work, \textbf{PANDA}, aims to address this gap by introducing a more adaptive augmentation strategy tailored for continual learning settings.

% -----------------------------------------------------------------------
\section{Methodology}
\label{sec:method}

\subsection{Preliminaries for Long-Tailed Continual Learning}
We assume the problem of supervised classification in the context of continual learning~\cite{CIL_survey}, where we encounter a series of N tasks ${T_1, T_2,...,T_N}$. Each task $T_k$ where $k \in N$ contains a disjoint set of images from a dataset paired with its corresponding labels. The overall dataset is structured to follow a long-tailed distribution, with the ordering of head and tail classes shuffled to better represent real-world scenarios. Based on this dataset, we construct data streams that we treat as distinct tasks The distribution is characterized by an exponential decay~\cite{cao2019learning}, parameterized by $\rho$ denotes the ratio between the most and least frequent classes. During training, we can only access the data from the current task $T_k$, where $C^k$ denotes the number of classes in the task $k$ and $n_j^k$ denotes the number of samples in class $j$ of task $k$.  As we are focusing on the exemplar-free setting, we don't have any storage systems for rehearsal and focus on learning the samples from the current task. Training samples in any task $k$, denoted as $\mathscr{X}_k=\{x^k_j, y^k_j | j \in \{1,2,...,C^k\}\}$ are i.i.d. samples drawn from the current distribution $D$ characterized by $\rho$, where $x^k_j$ are the images in task $k$ and $y^k_j$ are the corresponding labels. The goal of the continual learner is to classify all classes learned up to task $k$ denoted by $C^{1:k}$, which implies not only learning new task information, but also reducing catastrophic forgetting on previous tasks.

\subsection{Dual level imbalance}
\begin{figure}[!htbp]
    \centering
    \includegraphics[width=1\linewidth]{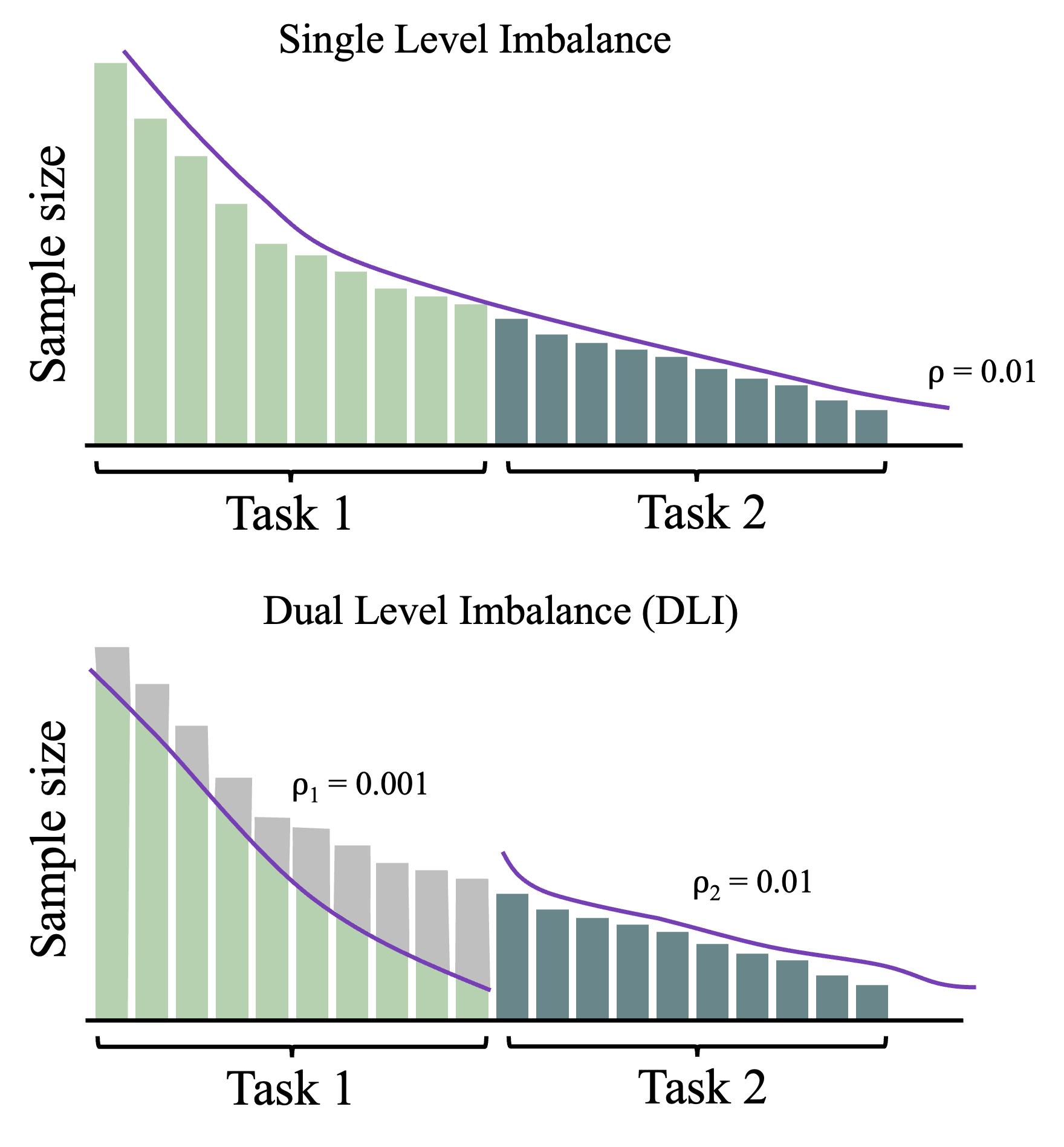}
    \caption{An illustrative figure for the Dual Level Imbalance (DLI) setting: the top image represents the conventional long-tailed CL scenario (SLI), while the bottom image demonstrates the DLI setup, where task-level imbalances are introduced in addition to the dataset-level imbalance. \textbf{NOTE: The order of tasks and classes in tasks are shuffled, and are not ordered from head to tail.}}
    \label{fig:DLI}
\end{figure}

In this section, we formally define the Dual-Level Imbalance (DLI) setup. Prior work on imbalanced continual learning~\cite{liu2022long, ocl_imbalance, sraghavan2024delta, APART, DAP} primarily focusing on the exponential decay of class samples (controlled by parameter $\rho$) and do not explicitly control how imbalance impacts individual tasks, we call this the Single-Level Imbalance (SLI) setting. Although SLI setting introduce varying levels of imbalance across the entire dataset, they fail to address task-specific imbalance. To bridge this gap, we propose a DLI setup as shown in Figure~\ref{fig:DLI} that incorporates both dataset-level (controlled by $\rho$) and task-level (controlled by $\rho^*$, $*$ denotes the task affected) imbalances. This approach amplifies inter-task imbalance (between different tasks) and can intensify intra-task imbalance, making the problem more realistic and challenging in the context of continual learning.

\begin{figure*}[h!]
    \centering
    \includegraphics[width=1\linewidth]{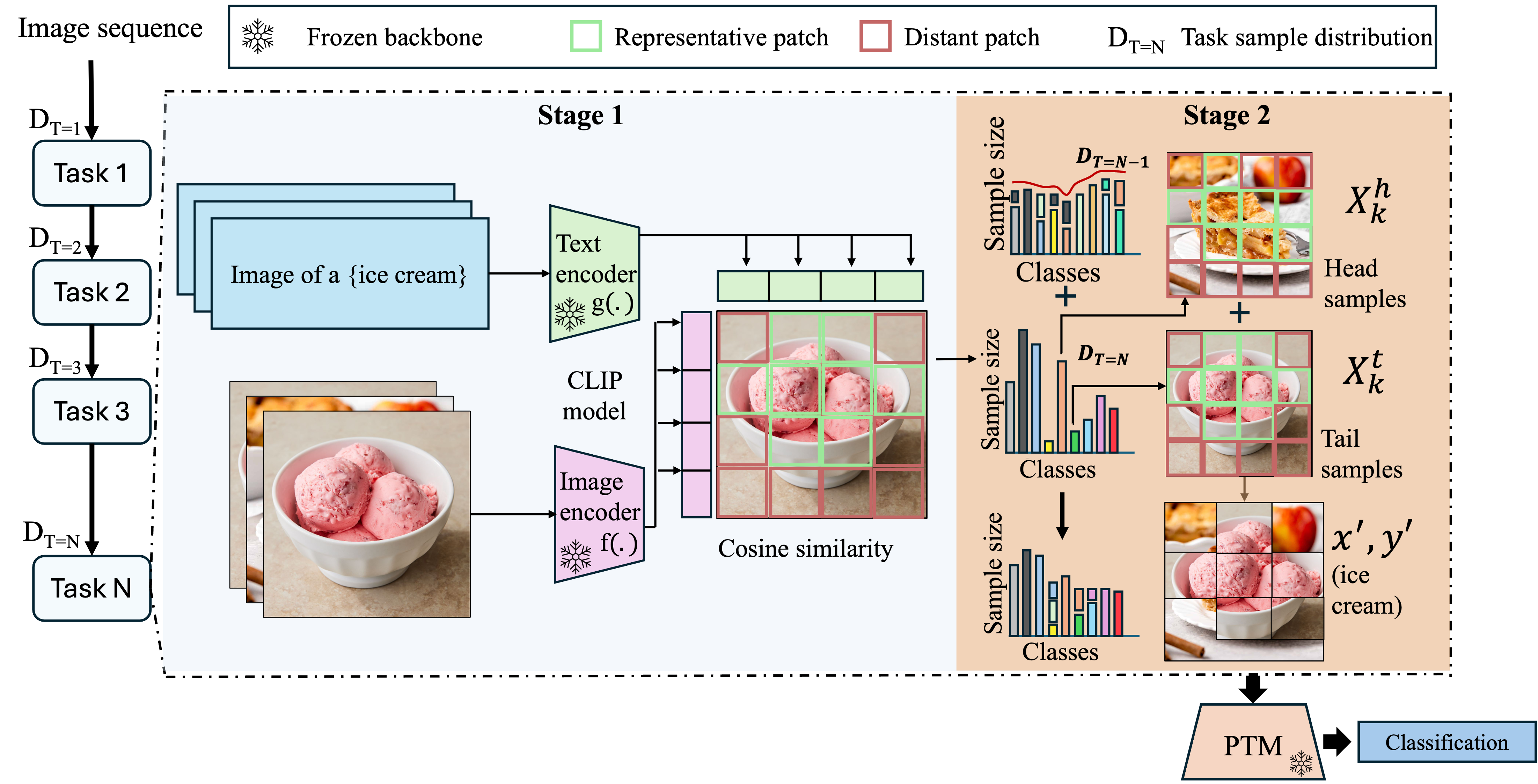}
    \caption{An overview of the PANDA framework, designed to improve Long-Tail Continual Learning through a training-free approach that contextually enriches tail-end samples by utilizing the distributions of head-class samples task-wise. We employ the frozen CLIP~\cite{CLIP} model to identify and extract the most relevant patches from head-class samples, transferring them to tail-class samples. Additionally, we incorporate prior task distributions to balance the current task, to mitigate inter-class and inter-task imbalances. The augmented samples are then fed into the Continual Learning pipeline.}
    \label{fig:main_overview}
\end{figure*}

\subsection{PANDA Framework}

%-----------------------------
We introduce PANDA, a training-free debiasing augmentation framework. The core objective of PANDA is to increase the effective number of training samples for tail classes (least frequent) by leveraging the rich contextual diversity found in head-class images (most frequent). Following the mechanism of vision transformers, each image is divided into $N \times N$ non-overlapping regions, referred to as patches. PANDA synthesizes new images by transferring the most semantically relevant patches, which are identified using a frozen CLIP~\cite{CLIP} encoder, from tail-class samples to the head-class samples. Images typically consist of an object of interest with background content that does not affect the classification context. By using this knowledge during patch swapping from head to tail classes, PANDA enriches the diversity of tail-class samples and improves the class distribution and reduces bias towards head class data.

% At any task $T_k$ with distribution $D_{T=k}$ the training samples are represented as $\mathscr{X}_k = \{ x_j^k, y_j^k | j\in\{ 1, 2, 3,...,C^k\} \}$.
% Essentially, this can be split into head and tail classes based on the sample frequency

At task \(T_k\) with sample distribution \(D_{T=k}\), the training set is  
\[
\mathscr{X}_k = \{(x_j^k,y_j^k)\mid j=1,\dots,C^k\}.
\]
We partition \(\mathscr{X}_k\) into head and tail classes according to sample frequency,  
\begin{align}
    \mathscr{X}_k &= \{ \mathscr{X}_k^h, \; \mathscr{X}_k^t \}, \\
    \text{with distributions} &= \;[D_{T=k}^h, \; D_{T=k}^t].
\end{align}

Let each head-class image be denoted as \(x^h \in \mathbb{R}^{H \times W \times C}\) with label \(y^h\) and each tail-class image as \(x^t \in \mathbb{R}^{H \times W \times C}\) with label \(y^t\). Our goal is to compose a new sample \((x', y')\) that balances the intra-task distribution and leverages prior-task distribution statistics to mitigate inter-task distribution shifts.

Using a frozen pretrained CLIP encoder, we partition each image \(x^h\) and \(x^t\) into \(N\) semantic patches, each annotated with positional encoding \(P_e\). To exploit CLIP’s joint language–vision alignment, we then convert the class label into a pseudo-sentence of the form:

\begin{equation}
    t = \textbf{Image of a \{label\}}
\end{equation}
We then compute text and image embeddings using the frozen CLIP encoders (Eqn~\ref{eq:features}). 

\begin{equation}\label{eq:features}
\begin{aligned}
    z_t &= g(t) && \text{(text features)} \\
    z_i &= f(P_i) && \text{(image patch features)}
\end{aligned}
\end{equation}
The cosine similarity between $z_t$ and each patch feature $z_i$ is computed as
\begin{equation}\label{eq:cos_sim}
    S_i = \frac{z_i \cdot z_t}{\|z_i\|\|z_t\|}, \quad i=1,\ldots,N.
\end{equation}

For each image, we select the top 
N/2 patches whose similarity scores exceed a confidence threshold of 0.45 (determined by experiments, overall range 0 to 1), thereby preventing cross‐class contamination and eliminating bad patch choices. We denote their indices by  
\[
i_h^* = \arg\!\top_{k=N/2} \{S_i^h\},
\quad
i_t^* = \arg\!\top_{k=N/2} \{S_i^t\}.
\]
We define binary masks \(M^h, M^t\in\{0,1\}^{H\times W}\) so that  
\[
M(u,v)=
\begin{cases}
1,&(u,v)\in\text{patches }i^*,\\
0,&\text{otherwise}.
\end{cases}
\]
Let \((M^h)'\) be the inverse of \(M^h\). We compose the new image as  
\begin{align}
x' &= (M^h)' \odot x^h \;+\; M^t \odot x^t,\\
y' &= y^t,
\end{align}
where \(\odot\) denotes element wise mask multiplication. This procedure grafts the tail object’s most semantic patches into the head image while preserving its original context.

To prevent overfitting to these synthetic compositions, we apply standard image augmentations including flip crop, color jitter and Gaussian blur, before passing \(x'\) to the learner. We iterate this augmentation until the average class counts for head and tail differ by at most $q$ samples, achieving  

\begin{align}
\mathscr{X}_k^h &= \mathscr{X}_k^h - M,\quad \\
\mathscr{X}_k^t &= \mathscr{X}_k^t + M,\quad \\
\bigl|\mathrm{Avg}[D_{T=k}^h] &- \mathrm{Avg}[D_{T=k}^t]\bigr| \le q.
\end{align}

\subsubsection{Adaptive Distribution Smoothening}

To align distributions across tasks, we maintain vectors of prior task maxima and minima and compute  
\begin{equation}\label{eq:intra_task_imb}
\mathrm{adjusted}_m
= \beta\,\mathrm{prior}_m + (1-\beta)\,\mathrm{current}_m,
\quad m\in\{\min,\max\}.
\end{equation}

The coefficient \(\beta\) modulates how strongly the previous task’s performance influences the current one, smoothing the transition from $(D_{T=k-1})$ to $(D_{T=k})$. When the performance of current task drops compared to the previous one, $\beta$ is lowered to enable rapid adaptation. An improvement in performance leads to an increase in $\beta$ to reinforce stability, and if performance is roughly unchanged, $\beta$ remains the same.
% ----------------------------------------------------------------------
\section{Experiments}
In this section, we provide a detailed overview of our experiments and the datasets utilized. We evaluate the performance of existing exemplar free continual learning methods with a PTM backbone on long-tailed datasets (Single and Dual level imbalances) both with and without the integration of our PANDA framework. Finally, we present ablation studies to demonstrate the contribution of each component in our proposed approach.
\subsection{Datasets}
We utilize two widely used publicly available datasets, CIFAR-100~\cite{Cifar100} and a 100-class subset of iNaturalist~\cite{inaturalist}. To introduce class imbalance, we generate long-tailed versions of CIFAR-100 using an exponential decay factor of $\rho = 0.01$, where $0 < \rho \leq 1$, which controls the ratio between the most and least frequent classes~\cite{sraghavan2024delta} at the dataset level. Task wise imbalance for DLI is controlled by $\rho^*$ where $*$ denotes the affected task. We ensure that the least frequent classes contain a minimum of three samples. The iNaturalist dataset inherently follows a long-tailed distribution representative of real-world scenarios, and we randomly select 100 classes using a fixed seed of 1993. All images were resized to an image resolution of $224 \times 224$

%%%%%%%%%%%%%%%%%%%%%%%%%%%%%%%%%%%%%%%%%%%%%%%%%%%%%%%%%%
%%%%%%%%%%%%% MAIN EXPS COMPARISON TABLE %%%%%%%%%%%%%%%%%
%%%%%%%%%%%%%%%%%%%%%%%%%%%%%%%%%%%%%%%%%%%%%%%%%%%%%%%%%%

\begin{table*}[ht]
\centering
\resizebox{1\textwidth}{!}{%
\begin{tabular}{l
                cc  % CIFAR100-LT (ρ=1)
                cc  % CIFAR100-LT (ρ=0.01)
                cc} % iNaturalist
\toprule
\textbf{Method}
  & \multicolumn{2}{c}{\textbf{CIFAR100-LT ($\rho=1$)}}
  & \multicolumn{2}{c}{\textbf{CIFAR100-LT ($\rho=0.01$)}}
  & \multicolumn{2}{c}{\textbf{iNaturalist (100 cls)}} \\
\cmidrule(lr){2-3} \cmidrule(lr){4-5} \cmidrule(lr){6-7}
  & \textbf{Avg Acc($\uparrow$)} & \textbf{Avg For($\downarrow$)}
  & \textbf{Avg Acc($\uparrow$)} & \textbf{Avg For($\downarrow$)}
  & \textbf{Avg Acc($\uparrow$)} & \textbf{Avg For($\downarrow$)} \\
\midrule
\multicolumn{7}{c}{\textbf{Prompt Methods}} \\
\midrule
L2P~\cite{L2P}
  & 89.23 & 6.41 
  & 73.34 & 7.87
  & 78.41 & 4.72 \\
CodaPrompt~\cite{codaprompt}
  & 91.30 & 5.26
  & 76.52 & 7.55
  & 83.85 & 4.58 \\
DualPrompt~\cite{dualprompt}
  & 87.36 & 10.38
  & 74.24 & 8.14
  & 81.39 & 10.69 \\
DAP~\cite{DAP}
  & 71.95    & 18.56
  & 62.98    & 15.13
  & 66.38    & 13.67 \\
\midrule
L2P + PANDA
  & --    & --
  & \textcolor{Cerulean}{\textbf{81.32}} ($\uparrow$ 7.98) & \textcolor{Cerulean}{\textbf{6.08}} ($\downarrow$ 1.79)
  & \textcolor{Cerulean}{\textbf{85.47}} ($\uparrow$ 7.06) & \textcolor{Cerulean}{\textbf{3.37}} ($\downarrow$ 1.35) \\
CodaPrompt + PANDA
  & --    & --
  & \textcolor{Cerulean}{\textbf{87.49}} ($\uparrow$ 2.94) & \textcolor{Cerulean}{\textbf{4.61}} ($\downarrow$ 2.94)
  & \textcolor{Cerulean}{\textbf{90.45}} ($\uparrow$ 6.60) & \textcolor{Cerulean}{\textbf{3.30}} ($\downarrow$ 1.28) \\
DualPrompt + PANDA
  & --    & --
  & \textcolor{Cerulean}{\textbf{81.00}} ($\uparrow$ 6.76) & \textcolor{Cerulean}{\textbf{7.38}} ($\downarrow$ 0.76)
  & \textcolor{Cerulean}{\textbf{85.44}} ($\uparrow$ 4.05) & \textcolor{Cerulean}{\textbf{9.44}} ($\downarrow$ 1.25) \\
DAP + PANDA
  & --    & --
  & \textcolor{Cerulean}{\textbf{67.17}} ($\uparrow$ 4.19) & \textcolor{Cerulean}{\textbf{12.63}} ($\downarrow$ 2.50)
  & \textcolor{Cerulean}{\textbf{69.15}} ($\uparrow$ 2.77) & \textcolor{Cerulean}{\textbf{10.88}} ($\downarrow$ 2.79) \\
\midrule
\multicolumn{7}{c}{\textbf{Other Methods}} \\
\midrule
SimpleCIL~\cite{simplecil_adam}
  & 82.40 & 7.33
  & 79.01 & 8.14
  & 89.90 & 4.48 \\
Adam w/ SSF~\cite{simplecil_adam}
  & 89.05 & 4.94
  & 86.55 & 4.63
  & 91.05 & 2.91 \\
RanPAC~\cite{ranpac}
  & 94.89 & 3.95
  & 90.35 & 5.22
  & 94.35 & 2.38 \\
EASE~\cite{ease}
  & 92.88 & 6.65
  & 89.94 & 6.76
  & 86.91 & 5.27 \\
CoFiMA~\cite{cofima}
  & 94.29    & 4.68
  & 93.05    & 5.57
  & \textcolor{Cerulean}{\textbf{94.55}} ($\uparrow$ 0.99)    & 3.88 \\
SLCA~\cite{slca}
  & 93.86    & 7.01
  & 91.73    & 6.73
  & 92.54   & 7.22 \\
FeCAM~\cite{fecam}
  & 91.15    & 4.57
  & 82.99    & 7.06
  & 87.87    & \textcolor{Cerulean}{\textbf{3.33}} ($\downarrow$ 1.11) \\
APART~\cite{APART}
  & 86.78    & 10.86
  & 81.94    & 13.42
  & 83.47    & 12.66 \\
APER~\cite{APER}
  & 90.93    & 5.24
  & 87.66    & 5.66
  & 92.22     & 3.17  \\
MOS~\cite{MOS}
  &94.26     & 3.53
  & 91.60     & 4.69
  &95.49     &2.77  \\
\midrule
SimpleCIL + PANDA
  & --    & --
  & \textcolor{Cerulean}{\textbf{80.20}} ($\uparrow$ 1.19) & \textcolor{Cerulean}{\textbf{7.98}} ($\downarrow$ 0.26)
  & \textcolor{Cerulean}{\textbf{91.92}} ($\uparrow$ 2.02)& \textcolor{Cerulean}{\textbf{4.44}} ($\downarrow$ 0.04) \\
Adam w/ SSF + PANDA
  & --    & --
  & \textcolor{Cerulean}{\textbf{88.08}} ($\uparrow$ 1.53) & \textcolor{Cerulean}{\textbf{4.32}} ($\downarrow$ 0.31)
  & \textcolor{Cerulean}{\textbf{92.61}} ($\uparrow$ 1.56) & \textcolor{Cerulean}{\textbf{2.38}} ($\downarrow$ 0.53) \\
RanPAC + PANDA
  & --    & --
  & \textcolor{Cerulean}{\textbf{91.91}} ($\uparrow$ 1.56) & \textcolor{Cerulean}{\textbf{4.38}} ($\downarrow$ 0.84)
  & \textcolor{Cerulean}{\textbf{95.70}} ($\uparrow$ 1.35) & \textcolor{Cerulean}{\textbf{1.97}} ($\downarrow$ 0.41) \\
EASE + PANDA
  & --    & --
  & \textcolor{Cerulean}{\textbf{91.97}} ($\uparrow$ 2.03) & \textcolor{Cerulean}{\textbf{6.65}} ($\downarrow$ 0.11)
  & \textcolor{Cerulean}{\textbf{92.45}} ($\uparrow$ 5.54) & \textcolor{Cerulean}{\textbf{5.25}} ($\downarrow$ 0.02) \\
CoFiMA + PANDA
  & --    & --
  & \textcolor{Cerulean}{\textbf{93.83}} ($\uparrow$ 0.78)    & \textcolor{Cerulean}{\textbf{4.91}} ($\downarrow$ 0.66)
  & 93.56    & \textcolor{Cerulean}{\textbf{2.98}} ($\downarrow$ 0.90) \\
SLCA + PANDA
  & --    & --
  & \textcolor{Cerulean}{\textbf{92.05}} ($\uparrow$ 0.32)   & \textcolor{Cerulean}{\textbf{6.23}}  ($\downarrow$ 0.50) 
  & \textcolor{Cerulean}{\textbf{93.27}}  ($\uparrow$ 0.73)    & \textcolor{Cerulean}{\textbf{4.58}} ($\downarrow$ 2.64) \\
FeCAM + PANDA
  & --    & --
  & \textcolor{Cerulean}{\textbf{86.48}} ($\uparrow$ 5.95)   & \textcolor{Cerulean}{\textbf{6.68}} ($\downarrow$ 0.38)
  & \textcolor{Cerulean}{\textbf{92.42}} ($\uparrow$ 4.55)   & 4.44 \\
APART + PANDA
  & --    & --
  & \textcolor{Cerulean}{\textbf{83.39}} ($\uparrow$ {1.45})    & \textcolor{Cerulean}{\textbf{11.48}} ($\downarrow$ {1.94})
  & \textcolor{Cerulean}{\textbf{85.91}} ($\uparrow$ {2.44})    & \textcolor{Cerulean}{\textbf{10.02}} ($\downarrow$ {2.64}) \\
APER + PANDA
  & --     & -- 
  & \textcolor{Cerulean}{\textbf{88.94}} ($\uparrow$ 1.28)     & \textcolor{Cerulean}{\textbf{5.26}} ($\downarrow$ 0.40)
   & \textcolor{Cerulean}{\textbf{92.42}} ($\uparrow$ 0.20)     & \textcolor{Cerulean}{\textbf{3.19}} ($\downarrow$ 0.02)\\
MOS + PANDA
  & --    & -- 
  & \textcolor{Cerulean}{\textbf{92.04}} ($\uparrow$ 0.80)     & \textcolor{Cerulean}{\textbf{4.48}} ($\downarrow$ 0.21)
  & \textcolor{Cerulean}{\textbf{95.85}} ($\uparrow$ 0.36)    & \textcolor{Cerulean}{\textbf{2.63}} ($\downarrow$ 0.14) \\
\bottomrule
\end{tabular}
}
\caption{Average top-1 accuracy in the \textbf{long-tailed} scenario (single-level imbalance) on CIFAR-100-LT (10 tasks) and iNaturalist (10 tasks) Best accuracy highlighted in \textcolor{Cerulean}{\textbf{boldface}}. NOTE: $\rho=1$ is the balanced case and PANDA framework has no added effect.}
\label{tab:overall_acc_table}
\end{table*}

\subsection{Implementation Details}
We adapt the publicly available PyTorch implementation of the PTM based CL algorithms from previous work~\cite{lamda_pilot, APART, DAP} for our experiments. In our PANDA framework, we utilize a frozen CLIP~\cite{CLIP} backbone. To ensure consistency, we split the datasets, construct long-tailed distributions, and set the random seed to 1993. The training batch size is set to 48, while the test batch size is 128. Optimizer and scheduler settings are inherited from each method individually, following the implementations in~\cite{lamda_pilot, APART, DAP}. Each experiment is repeated 10 times on a single NVIDIA A40 GPU, and we report the average accuracy and average forgetting in Table~\ref{tab:overall_acc_table} and the statical confidence is included in the supplementary material. In our PANDA framework, we select the top $N/2$ patches as the most representative based on the cosine similarity scores derived from comparing text and image embeddings.

\subsection{Evaluation Metrics}
In this paper, we use \textbf{Average Accuracy} and \textbf{Average Forgetting} to evaluate and compare the performance of different continual learning methods with PTMs. Average Accuracy metric quantifies overall performance on a balanced test set from previously encountered tasks. Given a total of $k$ tasks, let $a_{m,n}$ represent the model's performance on a held-out test set for task $n$ after being trained sequentially from task 1 to $m$.

\begin{equation}
    \text{Average Accuracy ($A_k$) =} \frac{1}{k}\sum_{i=1}^{k}a_{k,i}
\end{equation}

Average Forgetting for a task $k$ is calculated as the difference between its maximum performance obtained in the past and the current performance~\cite{comprehensive_CL_survey}
\begin{equation}
    \text{Average Forgetting ($F_{m, k}$) =} \max_{i \in \{1,\dots,k-1\}} \bigl(a_{i,m} - a_{k,m}\bigr), \quad \forall \, j < k
\end{equation}

%%%%%%%%%%%%%%%%%%%%%%%%%%%%%%%%%%%%%%%%%%%%%%%%%%%%%%%%%%
%%%%%%%%%%%%% DUAL IMBALANCE COMPARISON TABLE %%%%%%%%%%%%
%%%%%%%%%%%%%%%%%%%%%%%%%%%%%%%%%%%%%%%%%%%%%%%%%%%%%%%%%%

\begin{table*}[t!]
\centering
\resizebox{1\textwidth}{!}{%
\begin{tabular}{@{}l|cc|cc|cc@{}}
\toprule
\multirow{2}{*}{Method} &
  \multicolumn{2}{c|}{$\rho^*= 0.05$, $*=2$, $\rho = 0.01$} &
  \multicolumn{2}{c|}{$\rho^*= 0.05$, $*=3$, $\rho = 0.01$} &
  \multicolumn{2}{c}{$\rho^*= 0.05$, $*=4$, $\rho = 0.01$} \\
\cmidrule{2-3} \cmidrule{4-5} \cmidrule{6-7}
& Avg Acc(\%) & Avg For(\%) & Avg Acc(\%) & Avg For(\%) & Avg Acc(\%) & Avg For(\%) \\
\midrule
                   & \multicolumn{6}{c}{CIFAR100-LT (10 tasks)}                                          \\ 
\midrule
CodaPrompt~\cite{codaprompt}         & 84.27 & 4.67 & 82.88 & 5.03 & 82.29 & 4.80 \\
RanPAC~\cite{ranpac}             & 89.74 & 4.10 & 90.21 &  \textcolor{Cerulean}{\textbf{2.91}}($\downarrow$ 0.54) & 88.39 & 4.26 \\
% EASE~\cite{ease}               & 89.50 & 4.97 & -- & -- & -- & -- \\ 
MOS~\cite{MOS}                & \textcolor{Cerulean}{\textbf{93.54}} ($\uparrow$ 1.32) & 3.14 & 92.10 & 3.34 & 91.69 & 4.13 \\
CoFiMA~\cite{cofima}             & 93.97 & 4.02 & 92.18 & 4.66 & 90.39 & 5.17 \\ 
\midrule 
CodaPrompt + PANDA & \textcolor{Cerulean}{\textbf{89.44}} ($\uparrow$ 5.17) & \textcolor{Cerulean}{\textbf{3.62}} ($\downarrow$ 1.05) & \textcolor{Cerulean}{\textbf{87.77}} ($\uparrow$ 4.89) & \textcolor{Cerulean}{\textbf{3.46}} ($\downarrow$ 1.56) & \textcolor{Cerulean}{\textbf{85.35}} ($\uparrow$ 3.06) & \textcolor{Cerulean}{\textbf{4.62}} ($\downarrow$ 0.18) \\
RanPAC + PANDA     & \textcolor{Cerulean}{\textbf{90.89}} ($\uparrow$ 1.15) & \textcolor{Cerulean}{\textbf{3.84}} ($\downarrow$ 0.26) & \textcolor{Cerulean}{\textbf{92.65}} ($\uparrow$ 2.44) & 3.45  & \textcolor{Cerulean}{\textbf{90.05}} ($\uparrow$ 1.66) & \textcolor{Cerulean}{\textbf{3.93}} ($\downarrow$ 0.33) \\
% EASE + PANDA       & \textcolor{Cerulean}{\textbf{91.59}} ($\uparrow$ 2.09) & \textcolor{Cerulean}{\textbf{4.51}} ($\downarrow$ 0.46) & \textcolor{Cerulean}{\textbf{--}} ($\uparrow$ --) & \textcolor{Cerulean}{\textbf{--}} ($\downarrow$ --) & \textcolor{Cerulean}{\textbf{--}} ($\uparrow$ --) & \textcolor{Cerulean}{\textbf{--}} ($\downarrow$ --) \\ 
MOS + PANDA        & 92.22  & \textcolor{Cerulean}{\textbf{2.61}} ($\downarrow$ 0.53) & \textcolor{Cerulean}{\textbf{93.21}} ($\uparrow$ 1.11) & \textcolor{Cerulean}{\textbf{2.80}} ($\downarrow$ 0.54) & \textcolor{Cerulean}{\textbf{92.82}} ($\uparrow$ 1.13) & \textcolor{Cerulean}{\textbf{3.18}} ($\downarrow$ 0.95) \\
CoFiMA + PANDA        & \textcolor{Cerulean}{\textbf{94.38}} ($\uparrow$ 0.41 ) & \textcolor{Cerulean}{\textbf{3.27}} ($\downarrow$ 0.75) & \textcolor{Cerulean}{\textbf{93.25}} ($\uparrow$ 1.07) & \textcolor{Cerulean}{\textbf{3.86}} ($\downarrow$ 0.80) & \textcolor{Cerulean}{\textbf{92.05}} ($\uparrow$ 1.66) & \textcolor{Cerulean}{\textbf{4.82}} ($\downarrow$ 0.35) \\
\bottomrule
\end{tabular}
}
\caption{Average Accuracy (\%) and Forgetting (\%) on CIFAR100-LT with dual-level imbalance. \textit{$\rho$} indicates dataset level imbalance, \textit{$\rho^*$} indicates task level imbalance and \textit{$*$} indicates the task. The best results are in \textcolor{Cerulean}{\textbf{boldface}}}
\label{tab:dli_acc}
\end{table*}

\subsection{Results and Analysis}
\subsubsection{Conventional Long-Tailed setting}

In this section, we compare and discuss the performance of a wide range of PTM‑based exemplar‑free continual learning baselines against their counterparts integrated with our PANDA framework in the single and dual level long‑tailed setting. The evaluation is carried out on CIFAR-100~\cite{Cifar100}, and iNaturalist~\cite{inaturalist} with a Single Level Imbalance of $\rho=0.01$ and summarized in Table~\ref{tab:overall_acc_table}. It showcases that the performance of existing EFCL methods in the SLI long-tailed setting drops compared to the balanced setting ($\rho=1$), even on methods designed to tackle imbalanced data streams due to severe head class bias. 

In the case of \textbf{prompt-based methods} (L2P~\cite{L2P}, CodaPrompt~\cite{codaprompt}, DualPrompt~\cite{dualprompt}, DAP~\cite{DAP}), we attribute the reduced performance to a bias toward head-class samples. Unlike full fine-tuning, which updates the model's weights, prompt tuning optimizes only a small set of parameters. In highly imbalanced scenarios, this limited parameter adjustment may be insufficient to capture the diversity within tail classes, further exacerbating the performance gap. This limitation results in weak feature representations and poor classification performance. Despite DAP's dual anchor design tailored towards imbalanced distributions, head class samples still dominate its adaptation and retention. The tail class gradients are too weak to steer the general prompt, and the stabilizing anchor overlooks these tail class samples.

For \textbf{representation-based methods} (SimpleCIL, ADAM, RanPAC, EASE, MOS, APER, APART), except for EASE, the classifier is trained only on the first task, and in subsequent tasks, only the fully connected layer is updated using prototypes from the nearest mean classifier. In contrast, EASE introduces a new adapter for each task while keeping the PTM backbone frozen, allowing it to adapt to task-specific nuances during training. One reason for the lower performance in the long-tailed setting is the assumption that class distributions remain stable when updating the classifier using nearest mean prototypes. However, in continual learning with class imbalance, these distributions shift over time, making the prototypes less reliable and increasing the likelihood of misclassifications. 

Lastly, specialized approaches such as FeCAM~\cite{fecam}, SLCA~\cite{slca}, and CoFiMA~\cite{cofima} address continual learning from different perspectives. FeCAM leverages prototype-based classification using class centroids, CoFiMA ensembles model weights across tasks, and SLCA adapts learning rates and aligns classifiers to mitigate overfitting. While effective on balanced data, these methods end up optimizing on biased streams on frequent classes, which results in poorer representation of tail classes and thus lowered performance. Interestingly, in the iNaturalist dataset, we observe that CoFiMA without PANDA achieves higher accuracy than with PANDA integration. However, incorporating PANDA significantly reduces forgetting. For FeCAM, this is reversed. We attribute this behavior to the stability–plasticity trade-off. These methods are already operating near their performance upper bounds, thus improvements in forgetting lead to reductions in accuracy, and vice versa .

% By integrating our PANDA module, we simultaneously rebalance data \textbf{within tasks} by transferring patches from rare (tail) to frequent (head) classes. And \textbf{across tasks} by modulating a smoothing coefficient $\beta$ in response to recent performance trends, so that larger distribution shifts trigger stronger rebalancing and vice versa. This two‑level approach both boosts overall accuracy and curbs class bias and integrates seamlessly into existing PTM based EFCL methods.

\subsubsection{Dual-Level Imbalance (DLI) setting}
Next, we compare the results in the DLI setting proposed in our paper and present the performance in Table~\ref{tab:dli_acc} for the top four methods selected from Table~\ref{tab:overall_acc_table}. We notice that the existing methods struggle when the both, dataset-level and task-level imbalance is severe. We present 3 different cases with varying levels of imbalance (lower the $\rho$, more severe the imbalance). $\rho$ represents dataset-level imbalance and $\rho^*$ represents the task-level imbalance. By incorporating PANDA, we demonstrate that leveraging prior task distribution information alongside current task imbalances to stabilize the present task distribution and thereby reduce distribution shifts from previous tasks. PANDA helps us to balance the intra-task distributions and help smoothen shifts between tasks, which we attribute to our performance improvements.

%%%%%%%%%%%%%%%%%%%%%%%%%%%%%%%%%%%%%%%%%%%%%%%%%%%%%%%%%%
%%%%%%%%%%%%% LT AUGMENTATION COMPARISON TABLE %%%%%%%%%%%
%%%%%%%%%%%%%%%%%%%%%%%%%%%%%%%%%%%%%%%%%%%%%%%%%%%%%%%%%%

\begin{table*}[!htbp]
\centering
\resizebox{1\textwidth}{!}{%
\begin{tabular}{l|cccc}
\hline
 & \multicolumn{2}{c|}{Single level imbalance}             & \multicolumn{2}{c}{Dual level imbalance} \\ \hline
Augmentation technique &
  \multicolumn{2}{c|}{$\rho = 0.02$}  &
  \multicolumn{2}{c}{$\rho^* = 0.05, *=3, \rho = 0.02$}\\ \hline
 &
  \multicolumn{1}{c|}{Average Accuracy (\%)} &
  \multicolumn{1}{c|}{Average Forgetting(\%)} &
  \multicolumn{1}{c|}{Average Accuracy (\%)} &
  \multicolumn{1}{c}{Average Forgetting(\%)} \\ \hline
  % & \multicolumn{4}{c}{CIFAR100-LT (10 tasks)}                                                         \\ \hline
RanPAC & \multicolumn{1}{c|}{84.39}      & \multicolumn{1}{c|}{5.82}      & \multicolumn{1}{c|}{85.07}         & 5.97         \\
RanPAC + CutMix~\cite{cutmix}      & \multicolumn{1}{c|}{85.43}      & \multicolumn{1}{c|}{7.97}      & \multicolumn{1}{c|}{84.03}         & 6.77         \\
RanPAC + Mixup~\cite{mixup}        & \multicolumn{1}{c|}{81.33}      & \multicolumn{1}{c|}{8.03}      & \multicolumn{1}{c|}{77.29}         & 7.06          \\
RanPAC +Remix~\cite{remix}         & \multicolumn{1}{c|}{86.50}      & \multicolumn{1}{c|}{7.55}      & \multicolumn{1}{c|}{86.51}         & 5.73         \\
RanPAC +Con-CutMix~\cite{concutmix} & \multicolumn{1}{c|}{87.27}      & \multicolumn{1}{c|}{6.48}      & \multicolumn{1}{c|}{84.19}         & 6.01         \\ \hline
RanPAC + PANDA (Ours)                                     & \multicolumn{1}{c|}{\textcolor{Cerulean}{\textbf{90.31}}} & \multicolumn{1}{c|}{\textcolor{Cerulean}{\textbf{5.03}}} & \multicolumn{1}{c|}{\textcolor{Cerulean}{\textbf{90.08}}}    & \textcolor{Cerulean}{\textbf{4.52}}    \\ \hline
\end{tabular}
}
\caption{Average Accuracy (\%) and Average Forgetting (\%) on \textbf{CIFAR100-LT} with single and Dual-Level Imbalance compared with other LT augmentation techniques. \textit{$\rho$} indicates dataset level imbalance, \textit{$\rho^*$} indicates task level imbalance and \textit{$*$} indicates the task. The best results are in \textcolor{Cerulean}{\textbf{boldface}}}
\label{tab:comp_aug}
\end{table*}
% --------------------------------------------------------------------
\section{Ablation Studies}
\label{sec:ablation}

\subsubsection{Comparison against Long-Tailed Learning Augmentation methods}

In this section, we present experiments highlighting the effectiveness of various components of PANDA. Specifically, we investigate different imbalance scenarios under the proposed DLI setup and compare the performance of existing augmentation methods designed for long-tailed learning to PANDA. We report the average accuracy on RANPAC in Table~\ref{tab:comp_aug} for four popular augmentation techniques, namely, CutMIX~\cite{cutmix}, Mixup~\cite{mixup}, Remix~\cite{remix} and Contrastive CutMix~\cite{concutmix} designed for long-tailed learning, and compare them with our proposed PANDA framework. We add RanPAC as the baseline for comparison to analyze the performance of other augmentation techniques, in addition to our proposed PANDA framework. From Table~\ref{tab:comp_aug} we can see that the existing methods like CutMix and Mixup do not even reach the baseline performance. We attribute this primarily due to the algorithm, which doesn't take the distributions into the balancing of classes. Comparatively, Remix performs better than the baseline because Remix is designed to adjust the mixing ratio and region based on the content of the images. This adaptive behavior allows the model to see a more informative and balanced mix of features, leading to better performance. In the single-level imbalance scenario, Contrastive CutMix outperforms other augmentation techniques and comes close to PANDA by using contrastive loss to align the representations of augmented views, thereby enhancing feature learning. However, in the Dual-Level Imbalance (DLI) setting, significant distribution shifts across tasks hinder performance improvements, a challenge that our proposed method PANDA effectively overcomes, achieving the best accuracy and minimum forgetting.

\subsubsection{Comparison Against other masking methods}

\begin{figure}[!h]
    \centering
    \includegraphics[width=1\linewidth]{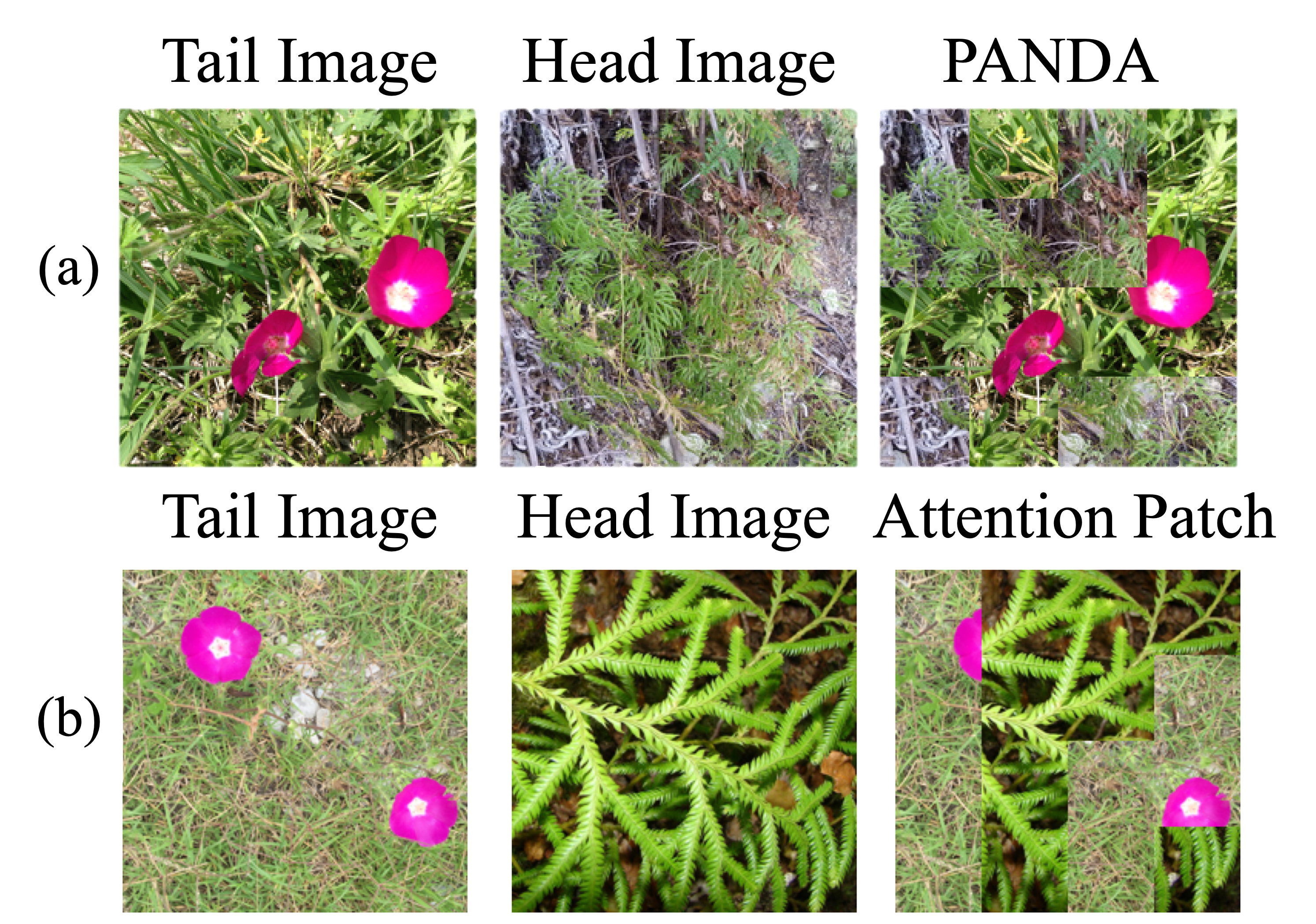}
    \caption{The figure illustrates the effect of (a) \textbf{TOP:} PANDA patching  and (b) \textbf{BOTTOM:} Attention based patching. Top-k patches from tail image are transferred to the head image.}
    \label{fig:abl-panda-attn-patch}
\end{figure}

In PANDA, we leverage a frozen CLIP encoder to isolate the most semantically salient patches for synthesizing augmented tail-class images. To further validate the ability of this CLIP based selection, we compare against \textbf{Attention Affinity Masking}. Inspired by the attention masking in modern methods like DinoV2~\cite{oquab2023dinov2}, we extract multi-head self-attention maps from a frozen vision transformer, average across heads, normalize into a soft affinity mask, and threshold the top N/2 patches by highest attention scores. As shown in Table~\ref{tab:affinityvspanda}, PANDA’s CLIP-similarity masks consistently outperform the attention affinity alternative. We attribute these gains to language-guided semantic alignment, which more effectively isolates representative regions in an image. In contrast, affinity masks capture noisy backgrounds, naive cropping lacks semantic focus. Beyond the quantitative results, we include qualitative comparisons of the original head and tail images alongside their augmented versions, PANDA patching in Figure~\ref{fig:abl-panda-attn-patch}(a) and Attention-Affinity patching in Figure~\ref{fig:abl-panda-attn-patch}(b). As illustrated in the two figures, the Attention–Affinity patching approach fails to capture the most representative regions of the head image, resulting in a patched output that blends features from both head and tail images and leads to class confusion. In contrast, Figure~\ref{fig:abl-panda-attn-patch}(a) shows that the representative regions of the tail image are completely patched into the head image, thereby avoiding any confusion between classes.

%%%%%%%%%%%%%%%%%%%%%%%%%%%%%%%%%%%%%%%%%%%%%%%%%%%%%%%%%%
%%%%%%%%%%%%% MASKING   COMPARISON TABLE %%%%%%%%%%%%%%%%%
%%%%%%%%%%%%%%%%%%%%%%%%%%%%%%%%%%%%%%%%%%%%%%%%%%%%%%%%%%
\begin{table}[!htbp]
\centering
\resizebox{1\linewidth}{!}{%
\begin{tabular}{@{}l|ll@{}}
\toprule
APART + Masking method           & Avg Acc (\%) ($\uparrow$) & Avg For (\%) ($\downarrow$)\\ \midrule
Baseline & 81.94 & 13.42 \\
Attention Affinity          & \textcolor{Red}{80.87} ($\downarrow$ 1.07)             & \textcolor{Red}{14.63} ($\uparrow$ 1.21)             \\
PANDA (\textit{ours}) & \textcolor{Cerulean}{\textbf{83.39}} ($\uparrow$ 1.45)            &  \textcolor{Cerulean}{\textbf{11.48}} ($\downarrow$ 1.94)               \\ \bottomrule
\end{tabular}
}
\caption{Performance comparison on Attention-Affinity Mask vs. Low Level Feature Transfer vs CLIP-Similarity Mask on 10 tasks, $\rho=0.01$ with CIFAR100-LT on the APART~\cite{APART}. \textcolor{red}{Decreased} and \textcolor{Cerulean}{improved} performance is highlighted. }
\label{tab:affinityvspanda}
\end{table}

\subsubsection{Resource usage}
Additionally, we detail the computational resources required including GPU memory usage and runtime for existing algorithms in Table~\ref{tab:resources}. We compare these metrics to those observed with the integration of PANDA. A modest resource increase substantially boosts continual learning performance by reducing bias, achieved by balancing distributions both within and across tasks.

%%%%%%%%%%%%%%%%%%%%%%%%%%%%%%%%%%%%%%%%%%%%%%%%%%%%%%%%%%
%%%%%%%%%%%%% GPU USAGE TABLE %%%%%%%%%%%%%%%%%%%%%%%%%%%%
%%%%%%%%%%%%%%%%%%%%%%%%%%%%%%%%%%%%%%%%%%%%%%%%%%%%%%%%%%

\begin{table}[!h]
\centering
\resizebox{1\columnwidth}{!}{%
\begin{tabular}{@{}ccc@{}}
\toprule
\multicolumn{3}{c}{iNaturalist (100 classes) - 10 tasks}                   \\ \midrule
\multicolumn{1}{c|}{}             & \multicolumn{1}{c|}{GPU Memory usage (MB)} & Running time (Hours) \\ \midrule
\multicolumn{1}{c|}{L2P}          &\multicolumn{1}{c|}{2994}                &\multicolumn{1}{c}{1.31}              \\
\multicolumn{1}{c|}{CodaPrompt}   &\multicolumn{1}{c|}{19234}                &\multicolumn{1}{c}{1.16}             \\
\multicolumn{1}{c|}{RanPAC}       &\multicolumn{1}{c|}{5052}                &\multicolumn{1}{c}{0.33}              \\
\multicolumn{1}{c|}{ADAM w/ SSF}  &\multicolumn{1}{c|}{9050}               &\multicolumn{1}{c}{0.61}          \\ 
\multicolumn{1}{c|}{CoFiMa}  &\multicolumn{1}{c|}{16488} & \multicolumn{1}{c}{1.20}         \\ 
\multicolumn{1}{c|}{SLCA}  & \multicolumn{1}{c|}{12220} & \multicolumn{1}{c}{0.94}        \\ \midrule
\multicolumn{1}{c|}{L2P + PANDA} &\multicolumn{1}{c|}{3282 (+288 MB $\uparrow$)}               & \multicolumn{1}{c}{1.59 (+0.28 Hr $\uparrow$)}           \\
\multicolumn{1}{c|}{CodaPrompt + PANDA} & \multicolumn{1}{c|}{19368 (+134 MB $\uparrow$)} & \multicolumn{1}{c}{1.33 (+0.17 Hr $\uparrow$)} \\
\multicolumn{1}{c|}{RanPAC + PANDA}      & \multicolumn{1}{c|}{5517 (+465 MB $\uparrow$)} & \multicolumn{1}{c}{0.42 (+0.09 Hr $\uparrow$)} \\
\multicolumn{1}{c|}{ADAM w/ SSF + PANDA} & \multicolumn{1}{c|}{9384 (+334 MB $\uparrow$)} & \multicolumn{1}{c}{1.07 (+0.46 Hr $\uparrow$)} \\ 
\multicolumn{1}{c|}{CoFiMa+PANDA}  &\multicolumn{1}{c|}{17066 (+ 578 MB $\uparrow$)} & \multicolumn{1}{c}{1.43 (+0.23 Hr $\uparrow$)}         \\ 
\multicolumn{1}{c|}{SLCA+PANDA}  &\multicolumn{1}{c|}{12612 (+392 MB $\uparrow$)} & \multicolumn{1}{c}{1.08 (+0.14 Hr $\uparrow$)}          \\ 
\bottomrule
\end{tabular}
}
\caption{The running time and memory usage for existing methods compared with addition of our PANDA augmentation framework on \textbf{iNaturalist subset} dataset (100 classes). We highlighted the increase in resources with our framework}
\label{tab:resources}
\end{table}
% -----------------------------------------------------------------
\section{Limitations}
While we provide a comprehensive analysis, experimentation, and ablation of our proposed framework, PANDA nonetheless has some limitations. The method depends on a pre-trained CLIP model for feature alignment and patch selection, making its performance inherently tied to the efficiency of the model.

\section{Conclusion}
\label{sec:conclusion}

This paper introduces Patch and Distribution Aware Augmentation (PANDA) for LT-EFCL, leveraging pre-trained models to address imbalances. PANDA is a training-free method that uses CLIP to extract tail-class patches and integrate them into head-class samples, which balances intra-task imbalances, smoothens inter-task distribution shifts, and reduces bias. We also formalize a Dual-Level Imbalance (DLI) setting for task-level imbalances. Extensive experiments show our method surpasses baselines to improve accuracy and mitigate forgetting.
% -----------------------------------------------------------------

\bibliography{aaai2026}

\clearpage
\section{Supplementary Material}
\begin{table*}[b]
\centering
\resizebox{1\textwidth}{!}{%
\begin{tabular}{l
                cc  % CIFAR100-LT (ρ=1)
                cc  % CIFAR100-LT (ρ=0.01)
                cc} % iNaturalist
\toprule
\textbf{Method}
  & \multicolumn{2}{c}{\textbf{CIFAR100-LT ($\rho=1$)}}
  & \multicolumn{2}{c}{\textbf{CIFAR100-LT ($\rho=0.01$)}}
  & \multicolumn{2}{c}{\textbf{iNaturalist (100 cls)}} \\
\cmidrule(lr){2-3} \cmidrule(lr){4-5} \cmidrule(lr){6-7}
  & \textbf{Avg Acc($\uparrow$)} & \textbf{Avg For($\downarrow$)}
  & \textbf{Avg Acc($\uparrow$)} & \textbf{Avg For($\downarrow$)}
  & \textbf{Avg Acc($\uparrow$)} & \textbf{Avg For($\downarrow$)} \\
\midrule
\multicolumn{7}{c}{\textbf{Prompt Methods}} \\
\midrule
L2P~\cite{L2P}
  & 89.23 $\pm$ 1.04 & 6.41 $\pm$ 0.33 
  & 73.34 $\pm$ 0.87 & 7.87 $\pm$ 0.39
  & 78.41 $\pm$ 0.63 & 4.72 $\pm$ 0.15 \\
CodaPrompt~\cite{codaprompt}
  & 91.30 $\pm$ 0.56 & 5.26 $\pm$ 0.09
  & 76.52 $\pm$ 0.87 & 7.55 $\pm$ 0.14
  & 83.85 $\pm$ 0.68 & 4.58 $\pm$ 0.03 \\
DualPrompt~\cite{dualprompt}
  & 87.36$\pm$1.02 & 10.38$\pm$ 0.25
  & 74.24$\pm$0.49 & 8.14$\pm$0.20
  & 81.39$\pm$0.16 & 10.69$\pm$0.19 \\
DAP~\cite{DAP}
  & 71.95$\pm$0.89    & 18.56$\pm$0.24
  & 62.98$\pm$0.60    & 15.13$\pm$0.67
  & 66.38$\pm$0.48    & 13.67$\pm$0.44 \\
\midrule
L2P + PANDA
  & --    & --
  & \textcolor{Cerulean}{\textbf{81.32 $\pm$1.09}} & \textcolor{Cerulean}{\textbf{6.08$\pm$0.40}} 
  & \textcolor{Cerulean}{\textbf{85.47$\pm$0.61}}  & \textcolor{Cerulean}{\textbf{3.37$\pm$0.13}} \\
CodaPrompt + PANDA
  & --    & --
  & \textcolor{Cerulean}{\textbf{87.49$\pm$0.48}} & \textcolor{Cerulean}{\textbf{4.61$\pm$0.19}} 
  & \textcolor{Cerulean}{\textbf{90.45$\pm$0.37}}& \textcolor{Cerulean}{\textbf{3.30$\pm$0.62}} \\
DualPrompt + PANDA
  & --    & --
  & \textcolor{Cerulean}{\textbf{81.00$\pm$0.18}} & \textcolor{Cerulean}{\textbf{7.38$\pm$0.06}} 
  & \textcolor{Cerulean}{\textbf{85.44$\pm$0.22}}  & \textcolor{Cerulean}{\textbf{9.44$\pm$0.11}}  \\
DAP + PANDA
  & --    & --
  & \textcolor{Cerulean}{\textbf{67.17$\pm$0.66}}  & \textcolor{Cerulean}{\textbf{12.63$\pm$0.17}} 
  & \textcolor{Cerulean}{\textbf{69.15$\pm$0.49}}  & \textcolor{Cerulean}{\textbf{10.88$\pm$0.08}}  \\
\midrule
\multicolumn{7}{c}{\textbf{Other Methods}} \\
\midrule
SimpleCIL~\cite{simplecil_adam}
  & 82.40$\pm$0.43 & 7.33$\pm$0.59
  & 79.01$\pm$0.61 & 8.14$\pm$0.32
  & 89.90$\pm$0.54 & 4.48$\pm$0.04 \\
Adam w/ SSF~\cite{simplecil_adam}
  & 89.05$\pm$0.33 & 4.94$\pm$0.45
  & 86.55$\pm$0.60 & 4.63$\pm$0.23
  & 91.05$\pm$0.42 & 2.91$\pm$0.65 \\
RanPAC~\cite{ranpac}
  & 94.89$\pm$0.27 & 3.95$\pm$0.60
  & 90.35$\pm$0.44 & 5.22$\pm$0.20
  & 94.35$\pm$0.72 & 2.38$\pm$0.33 \\
EASE~\cite{ease}
  & 92.88$\pm$0.68 & 6.65$\pm$0.37
  & 89.94$\pm$0.54 & 6.76$\pm$0.22
  & 86.91$\pm$0.32 & 5.27$\pm$0.48 \\
CoFiMA~\cite{cofima}
  & 94.29$\pm$0.30    & 4.68$\pm$0.66
  & 93.05$\pm$0.63    & 5.57$\pm$0.58
  & \textcolor{Cerulean}{\textbf{94.55$\pm$0.74}}   & 3.88$\pm$0.26 \\
SLCA~\cite{slca}
  & 93.86$\pm$0.64    & 7.01$\pm$0.46
  & 91.73$\pm$0.14    & 6.73$\pm$0.03
  & 92.54$\pm$0.08  & 7.22$\pm$0.12 \\
FeCAM~\cite{fecam}
  & 91.15$\pm$0.72    & 4.57$\pm$0.39
  & 82.99$\pm$0.53   & 7.06$\pm$0.19
  & 87.87$\pm$0.59    & \textcolor{Cerulean}{\textbf{3.33$\pm$0.30}} \\
APART~\cite{APART}
  & 86.78$\pm$0.67    & 10.86$\pm$0.60
  & 81.94$\pm$0.70    & 13.42$\pm$0.58
  & 83.47$\pm$0.32    & 12.66$\pm$0.31 \\
APER~\cite{APER}
  & 90.93$\pm$0.55    & 5.24$\pm$0.35
  & 87.66$\pm$0.49    & 5.66$\pm$0.44
  & 92.22$\pm$0.02     & 3.17 $\pm$0.01  \\
MOS~\cite{MOS}
  &94.26$\pm$0.40     & 3.53$\pm$0.33
  & 91.60$\pm$0.57     & 4.69$\pm$0.02
  &95.49$\pm$0.29     &2.77$\pm$0.06  \\
% DUCT~\cite{duct}
%   &     & 
%   &     & 
%   &     &  \\
\midrule
SimpleCIL + PANDA
  & --    & --
  & \textcolor{Cerulean}{\textbf{80.20$\pm$0.22}} & \textcolor{Cerulean}{\textbf{7.98$\pm$0.05}}
  & \textcolor{Cerulean}{\textbf{91.92$\pm$0.06}}& \textcolor{Cerulean}{\textbf{4.44$\pm$0.04}}\\
Adam w/ SSF + PANDA
  & --    & --
  & \textcolor{Cerulean}{\textbf{88.08$\pm$0.09}}& \textcolor{Cerulean}{\textbf{4.32$\pm$0.07}}
  & \textcolor{Cerulean}{\textbf{92.61$\pm$0.11}} & \textcolor{Cerulean}{\textbf{2.38$\pm$0.13}} \\
RanPAC + PANDA
  & --    & --
  & \textcolor{Cerulean}{\textbf{91.91$\pm$0.26}} & \textcolor{Cerulean}{\textbf{4.38$\pm$0.11}} 
  & \textcolor{Cerulean}{\textbf{95.70$\pm$0.21}} & \textcolor{Cerulean}{\textbf{1.97$\pm$0.21}}  \\
EASE + PANDA
  & --    & --
  & \textcolor{Cerulean}{\textbf{91.97$\pm$0.07}}  & \textcolor{Cerulean}{\textbf{6.65$\pm$0.08}} 
  & \textcolor{Cerulean}{\textbf{92.45$\pm$0.16}}  & \textcolor{Cerulean}{\textbf{5.25$\pm$0.10}}  \\
CoFiMA + PANDA
  & --    & --
  & \textcolor{Cerulean}{\textbf{93.83$\pm$0.17}}    & \textcolor{Cerulean}{\textbf{4.91$\pm$0.03}}
  & 93.56$\pm$ 0.16    & \textcolor{Cerulean}{\textbf{2.98$\pm$ 0.17}} \\
SLCA + PANDA
  & --    & --
  & \textcolor{Cerulean}{\textbf{92.05$\pm$0.04}}   & \textcolor{Cerulean}{\textbf{6.23$\pm$0.11}}  
  & \textcolor{Cerulean}{\textbf{93.27$\pm$0.08}}     & \textcolor{Cerulean}{\textbf{4.58$\pm$0.13}}  \\
FeCAM + PANDA
  & --    & --
  & \textcolor{Cerulean}{\textbf{86.48$\pm$0.31}} & \textcolor{Cerulean}{\textbf{6.68$\pm$0.07}} 
  & \textcolor{Cerulean}{\textbf{92.42$\pm$0.15}}  & 4.44$\pm$0.26 \\
APART + PANDA
  & --    & --
  & \textcolor{Cerulean}{\textbf{83.39$\pm$0.16}}    & \textcolor{Cerulean}{\textbf{11.48$\pm$0.09}}
  & \textcolor{Cerulean}{\textbf{85.91$\pm$0.11}}    & \textcolor{Cerulean}{\textbf{10.02$\pm$0.16}}  \\
APER + PANDA
  & --     & -- 
  & \textcolor{Cerulean}{\textbf{88.94$\pm$0.15}}      & \textcolor{Cerulean}{\textbf{5.26$\pm$0.02}} 
   & \textcolor{Cerulean}{\textbf{92.42 $\pm$0.06}}     & \textcolor{Cerulean}{\textbf{3.19 $\pm$0.07}} \\
MOS + PANDA
  & --    & -- 
  & \textcolor{Cerulean}{\textbf{92.04$\pm$0.02}}     & \textcolor{Cerulean}{\textbf{4.48$\pm$0.12}}
  & \textcolor{Cerulean}{\textbf{95.85$\pm$0.06}}     & \textcolor{Cerulean}{\textbf{2.63$\pm$0.11}}  \\
% DUCT + PANDA
%   &     & 
%   &     & 
%   &     &  \\
% \midrule
% Offline training
%   & 96.08 & --
%   & 94.95 & --
%   & 96.90 & -- \\
\bottomrule
\end{tabular}
}
\caption{Average top-1 accuracy in the \textbf{long-tailed} scenario (single-level imbalance) on CIFAR-100-LT and iNaturalist. Best accuracy highlighted in \textcolor{Cerulean}{\textbf{boldface}}. NOTE: $\rho=1$ is the balanced case and PANDA framework has no added effect.}
\label{tab:overall_acc_table_suppl}
\end{table*}
%%%%%%%%%%%%%%%%%%%%%%%%%%%%%%%%%%%%%%%%%%%%%%%%%%%%%%%%%%
%%%%%%%%%%%%% BETA COMPARISON TABLE %%%%%%%%%%%%%%%%%%%%%%
%%%%%%%%%%%%%%%%%%%%%%%%%%%%%%%%%%%%%%%%%%%%%%%%%%%%%%%%%%
\subsection{Adaptive Beta comparison}
In the Methodology Section of our main paper we introduced “performance‑based adaptivity” in which the smoothing coefficient $\beta$ is adjusted according to recent task accuracy trends to balance stability and plasticity. Here, we compare this approach with two alternative adjustment schemes. The first, implemented in our main paper, \textbf{task‑progression adaptivity}, begins from a base $\beta$ and incrementally increases it as tasks proceed, reflecting the intuition that later tasks can rely more heavily on accumulated knowledge, up to a fixed upper bound. The second, \textbf{distribution‑change adaptivity}, modifies $\beta$ purely based on the magnitude of the shift in class distribution between successive tasks. Large shifts lead to a substantial reduction in $\beta$ (favoring rapid adaptation), moderate shifts yield a moderate decrease, and minimal shifts cause a modest increase (increasing stability). And lastly, we add a naive \textbf{Task Progress} setup where the beta value increases with the task index reflecting the intuition that the system should start out more plastic and become progressively stable as tasks advance. We compare these three adaptive tuning schemes to provide a comprehensive evaluation of the stability–plasticity trade-offs.
\subsection{Additional performance insights}

In our main paper in the experiments section we had included a table to consolidate the performance metrics including average accuracy and average forgetting. We include the statistical significance of these methods in Table~\ref{tab:overall_acc_table_suppl} for more detailed insights regarding performance improvements showcased.

%————————————————————————————————————————
\begin{algorithm}[!htbp]
\caption{Adaptive Beta Computation: Performance Based}
\label{alg:performance_based}
\begin{algorithmic}[1]
  \Require performanceHistory,\,baseBeta
  \Ensure $\beta$
  \If{$|\mathit{performanceHistory}| < 2$}
    \State \Return $baseBeta$
  \EndIf
  \State recentPerf $\gets \mathit{performanceHistory}[-2\,:\,]$
  \If{$recentPerf[1] < recentPerf[0] - 0.05$}
    \State \Return $\max\bigl(baseBeta - 0.15,\;0.5\bigr)$
  \ElsIf{$recentPerf[1] > recentPerf[0] + 0.02$}
    \State \Return $\min\bigl(baseBeta + 0.1,\;0.9\bigr)$
  \Else
    \State \Return $baseBeta$
  \EndIf
\end{algorithmic}
\end{algorithm}
%————————————————————————————————————————
\begin{algorithm}[!htbp]
\caption{Adaptive Beta: Task Progress Strategy}
\label{alg:task_progress}
\begin{algorithmic}[1]
  \Require $baseBeta > 0$
  \Ensure $\beta$
  \If{taskNum is None or $taskNum < 0$}
    \State \Return $baseBeta$
  \EndIf
  \State $progress \gets \min\bigl(taskNum/10,\;1.0\bigr)$
  \State $adaptiveBeta \gets baseBeta + 0.3 \times progress$
  \State \Return $\min(adaptiveBeta,\;0.95)$
\end{algorithmic}
\end{algorithm}

%————————————————————————————————————————
\begin{algorithm}[!htbp]
\caption{Adaptive Beta Computation: Distribution Change}
\label{alg:distribution_change}
\begin{algorithmic}[1]
  \Require distributionChanges,\,baseBeta
  \Ensure $\beta$
  \If{$|\mathit{distributionChanges}| = 0$}
    \State \Return $baseBeta$
  \EndIf
  \State recentChange $\gets \mathit{distributionChanges}[-1]$
  \If{$recentChange > 0.5$}
    \State \Return $\max\bigl(baseBeta - 0.2,\;0.5\bigr)$
  \ElsIf{$recentChange > 0.2$}
    \State \Return $\max\bigl(baseBeta - 0.1,\;0.6\bigr)$
  \Else
    \State \Return $\min\bigl(baseBeta + 0.1,\;0.9\bigr)$
  \EndIf
\end{algorithmic}
\end{algorithm}

\end{document}